\documentclass{article}

\usepackage[preprint, nonatbib]{neurips_2023}

\usepackage[utf8]{inputenc} 
\usepackage[T1]{fontenc}    
\usepackage[hyphens]{url}            
\usepackage{hyperref}       
\usepackage{booktabs}       
\usepackage{nicefrac}       
\usepackage{microtype}      
\usepackage{xcolor}         

\usepackage{graphicx}
\usepackage{subfigure}
\usepackage{booktabs} 

\usepackage{amsmath,amssymb,amsfonts}
\usepackage{mathtools}  
\usepackage{balance}
\usepackage{color}
\usepackage{enumitem}

\usepackage{graphicx}
\usepackage[font={bf,footnotesize},labelsep=period]{caption}
\usepackage{makecell, multirow}
\usepackage{times}
\usepackage{hyperref}

\usepackage{lipsum}
\usepackage{outlines}
\usepackage{soul}
\usepackage{algorithm}
\usepackage{algorithmic}
\usepackage{multibib}
\newcites{Supp}{References}

\newcommand{\DEL}[1]{}

\title{Breaking MLPerf Training: A Case Study on Optimizing BERT}

\author{%
  Yongdeok Kim$^{1}$ \quad Jaehyung Ahn$^{1}$ \quad Myeongwoo Kim$^{1}$ \quad Changin Choi$^{1}$\\
  \textbf{Heejae Kim}$^{1}$ \quad \textbf{Narankhuu Tuvshinjargal}$^{1}$ \quad \textbf{Seungwon Lee}$^{1}$ \quad \textbf{Yanzi Zhang}$^{2}$\\
  \textbf{Yuan Pei}$^{2}$ \quad \textbf{Xiongzhan Linghu}$^{2}$ \quad \textbf{Jingkun Ma}$^{2}$ \quad \textbf{Lin Chen}$^{2}$ \\
  \textbf{Yuehua Dai}$^{2}$ \quad \textbf{Sungjoo Yoo}$^{3}$ \\
  $^{1}$Samsung Advanced Institute of Technology \quad $^{2}$Samsung R\&D Institute China Xian\\
  $^{3}$Seoul National University \\
  \texttt{\{yd.mlg.kim,jh91.ahn,k.myeong-woo,ci2015.choi,} \\ \texttt{heejaeee.kim,nate.tuvshin,seungw.lee,yanzi.zhang,} \\
  \texttt{yuan.pei,xz.linghu,jingkun.ma,lin81.chen,yuehua.dai\}@samsung.com} \\
  \texttt{sungjoo.yoo@gmail.com} \\
}

\begin{document}

\maketitle

\begin{abstract}
Speeding up the large-scale distributed training is challenging in that it requires improving various components
of training including load balancing, communication, optimizers, etc. We present novel approaches for fast
large-scale training of BERT model which individually ameliorates each component thereby leading to a new
level of BERT training performance. Load balancing is imperative in distributed BERT training since its training
datasets are characterized by samples with various lengths. Communication cost, which is proportional to the
scale of distributed training, needs to be hidden by useful computation. In addition, the optimizers, e.g., ADAM,
LAMB, etc., need to be carefully re-evaluated in the context of large-scale distributed training. We propose two
new ideas, (1) local presorting based on dataset stratification for load balancing and (2) bucket-wise gradient
clipping before allreduce which allows us to benefit from the overlap of gradient computation and synchronization
as well as the fast training of gradient clipping before allreduce. We also re-evaluate existing optimizers via
hyperparameter optimization and utilize ADAM, which also contributes to fast training via larger batches than
existing methods. Our proposed methods, all combined, give the fastest MLPerf BERT training of 25.1 (22.3)
seconds on 1,024 NVIDIA A100 GPUs, which is 1.33$\times$ (1.13$\times$) and 1.57$\times$ faster than the other top two (one)
submissions to MLPerf v1.1 (v2.0).
Our implementation and evaluation results are available at MLPerf v1.1$\sim$v2.1.


\end{abstract}

\section{Introduction}
\label{sec:intro}

Although large-scale data-parallel training has been actively
studied in recent years, there is still a large room for
improvements in making best use of extremely large-scale
GPU resources. 
Especially, due to the ever-increasing training cost, it is imperative to improve its efficiency, i.e., faster training. Load balancing and
communication/computation overlapping are two most important
factors which affect efficiency. We investigate these two
issues in the training of the representative model, BERT
on a large-scale GPU training system consisting of 1,024
NVIDIA A100 GPUs with 100 GB/s network bandwidth.

Load balancing of large-scale data-parallel training of language
models is challenging due to the fact that NLP
datasets, used in BERT pre-training, are characterized by the
diversity in input data size. It is mainly because the training
datasets are obtained by collecting texts from various
sources, e.g., web pages, books, news articles, etc. and these
texts tend to exhibit wide distributions in terms of sequence
length. Such a large variation in sequence lengths can incur
a significant load imbalance across GPUs thereby degrading
the training speed. For instance, a BERT model of sequence
length 512 wastes almost half the computation budget due
to padded tokens to short sentences in the Wikipedia dataset~\cite{krell2022efficient}.
Recently, a few studies have been presented to address this problem.
In \cite{web:mlperfv11nvidia}, the training data are first sorted in terms of sequence length
and balanced batches are formed by utilizing the sorted sequences.
Krell \textit{et al.}~\cite{krell2022efficient} propose a novel packing method,
which addresses this problem and shows that a maximum 2×
speedup is achievable. These existing methods
show a potential of improving training speed through better
load balancing. However, each incurs new overhead, e.g.,
allgather and sorting the training dataset~\cite{web:mlperfv11nvidia}, or
loses advantages in case of large batches, e.g., the benefit of CUDA graph cannot be exploited in local batches larger than 6 due to the lack of memory.
Furthermore, the packing method does not preserve the order of sentences,
so it impossible to use the packing method for tasks such as question-answer,
where the order of the sentences is important.

In standard data-parallel training, overlapping computation
and communication between compute devices is typically
realized by overlapping backward pass (i.e., gradient computation
on individual GPUs in parallel) and gradient synchronization
(i.e., allreduce via inter-GPU communication).
The communication and computation is overlapped in a fine-grained
manner. Specifically, a bucket of gradients is communicated
across GPUs while the next bucket of gradients
is being computed on each GPU. To further improve training efficiency, i.e., via fewer iterations of training, we investigated
the possibility of adopting sophisticated training
methods in the standard data-parallel training. Especially,
although gradient clipping before allreduce can accelerate
training, it has not been adopted in the data-parallel training
since it cannot benefit from the communication/computation
overlapping due to its requirement of performing gradient
clipping (involving inter-GPU communication) before
allreduce and the resulting serialization of computations
(gradient computation and clipping) and synchronization
(allreduce).

In this paper, we address two problems of load balancing
and overlapping and propose novel ideas for improving the
training speed of large-scale data-parallel BERT training.
In addition, we revisit the selection of optimizers in the
context of large-scale distributed BERT training and select
ADAM as the optimizer, enabling us to utilize larger
batches than the other optimizers while finally contributing
to fast training.

Our contributions are summarized as follows:
\begin{itemize}[itemsep=0mm]
    \item We investigate the characteristics of training datasets in NLP training and propose a novel and low-cost method of forming balanced batches based on stratification and local presorting.
    \item In order to address the problem of no overlapping between communication and computation in gradient clipping before allreduce, we propose performing gradient clipping in a bucket-wise manner, which is critical for us to achieve the SOTA training speed in BERT training.
    \item We also demonstrate the conventional ADAM is still effective in large-scale BERT training under hyperparameter optimization and can contribute to fast training with large batches.
\end{itemize}
\section{Background}
\label{sec:background}

\subsection{Data Parallelism}
Data parallelism plays a critical role in training
models with large parameters and/or large training data on
large-scale compute devices, e.g., GPUs.
In data parallelism, whether it is adopted in a purely data-parallel setting or used together with other types of parallelism, e.g., model/pipeline parallelism~\cite{web:deepspeed}, each GPU first stores its
own copy of the entire model in the beginning of training (in a purely data-parallel setting of small model training) or prepares the model parameters of each layer, on the fly, via allgather under ZeRO~\cite{web:deepspeed} (for large model training). Each GPU performs forward and backward passes of
training. The local gradients obtained in the backward pass
are exchanged across GPUs in order for each GPU to update
its copy of the entire model. There have been many studies
to improve the performance of data-parallel training: synchronous
(i.e., weight update in every iteration of training)
vs. asynchronous weight updates, parameter server vs. allreduce,
gradient compression, etc.~\cite{tang2020communicationefficient}. These
days, thanks to two key innovations to be briefly explained
below, a typical choice is synchronous allreduce without
gradient compression when adequate network bandwidth is
available~\cite{agarwal2022utility}.

\textbf{Allreduce~~~} There have been various optimizations of allreduce
such as ring-reduce~\cite{barnett1994interprocessor},
tree-reduce~\cite{sanders2009two}, recursive doubling~\cite{ueno2019exhaustive}, etc.
The optimizations are tightly coupled with the
underlying communication networks (topology and bandwidth),
which makes it difficult to optimize allreduce on a
given large-scale training platform consisting of thousands
of GPUs connected by a complicated network. However,
recent solutions such as NCCL from NVIDIA dynamically
selects suitable options, e.g., between double binary-tree-reduce
and ring-reduce, considering the number of GPUs,
and the topology, bandwidth and unit size of communication.

\textbf{Communication and computation overlap~~~} In the backward
pass of training, while the current gradients are being
computed, the previously calculated gradients can be synchronized,
i.e., allreduce can be performed on them across
GPUs in order for each GPU to obtain the final updates
to the model parameters (copied across the GPUs). Such a
capability of hiding communication with computation is one
of the key requirements for efficient data-parallel training.
The collective communication of allreduce is often realized
at the granularity of bucket (e.g., 25 MB by default in PyTorch)
in order to amortize the overhead of calling allreduce
collectives. It is crucial to realize efficient allreduce collectives
by dynamically ordering buckets while tracing the
backward order using autograd hooks and updating parameters
according to bucket mapping~\cite{li2020pytorch, sergeev2018horovod}.

\subsection{Large Batch Training}
Large batch training, which is required to make the best use
of large-scale compute resources, is known to suffer from
lack of generalization ability.
Krizhevsky proposes a simple heuristics that increases the
learning rate in proportion to the batch size and this strategy works up to a certain
batch size~\cite{krizhevsky2014weird}.
You \textit{et al.} propose a layer-wise adaptive rate
scaling method that scales parameter updates
by the $L_{2}$ norm of parameters~\cite{you2017large}.
You \textit{et al.} also propose LAMB method that combines the
layer-wise adaptive rate scaling with ADAM, which enables
large batch training for BERT and ResNet50 models~\cite{you2020large}.
It is also reported that, under sufficient
hyperparameter optimizations, the standard optimizers like
ADAM can offer even better results in large batch training.
In this work, as explained later, we also conducted
hyperparameter optimizations and found ADAM works best~\cite{you2020large}.
Thus, all the key results of our experiments were obtained
using ADAM optimizer.

\subsection{Gradient Clipping}
Gradient clipping, which is to scale the gradient if it gets
too large, is widely adopted for improving
training. In this paper, we apply clipping-by-norm
which is defined as follows. If $\| \mathbf{g} \| \geq c$ then
\begin{equation*}
	\mathbf{g} \leftarrow c \cdot \mathbf{g} / \| \mathbf{g} \|, 
\end{equation*}
where threshold c is typically set to 1 and $\| \mathbf{g} \|$ is the norm of
the gradients $\mathbf{g}$. The gradient scale is usually applied to the
entire gradients of a layer. Thus, it can be computed when
all the gradients are calculated, which means clipping needs
to be applied before or after allreduce. Although clipping
before allreduce proves more effective than clipping after
allreduce, it is not usually adopted since we cannot overlap
gradient computation and synchronization in gradient
clipping before allreduce. In this paper, we propose a novel
method to enable both gradient clipping before allreduce
and communication/synchronization overlapping.

\subsection{BERT Training in MLPerf Benchmark}

BERT training consists of two phases: (1) training from a random initialization with short sequences (at lengths of $<$128) to a certain accuracy and (2) training with longer sequences ($<$512) from the same checkpoint to a target accuracy. MLPerf benchmark adopts only the second phase since the training from the same checkpoint to a certain target accuracy level (72.0\%) enables fair and effective comparisons. Even though our experiments are performed only for the MLPerf benchmark for a few seconds, our methods can be applied to the entire training process and thus make a high contribution to training efficiency.
\cite{arxiv:2019:mlperf}

\section{Motivation}
\label{sec:motiv}

\subsection{Irregular Sequence Length in NLP Datasets}

The sequence length distribution of NLP datasets is characterized by a skewed
distribution. For example, in the Wikipedia dataset~\cite{wikidump},
the percentages of data with sequence lengths of 1 to 128, 129 to 256,
257 to 348, and 349 to 512 are 37.3\%, 19.7\%, 11.7\%, and 31.4\%, respectively.
\DEL{Table~\ref{tab:nlp} shows that the sequence length distribution of the
Wikipedia dataset (\cite{wikidump}) exhibits a skewed distribution
with dual modes.} It is
reported that SQUAD~\cite{rajpurkar2016squad} and GLUE
\cite{wang-etal-2018-glue} also have similar skewed distributions~\cite{krell2022efficient}.
The conventional method of handling irregular sequence
length in batch-based training is zero padding.
\DEL{For instance, the maximum sequence length is set to 512 in the
BERT pre-training with the Wikipedia dataset. }In such a
case, if a simple zero padding is adopted, approximately
50\% of GEMM computation is wasted due to matrix multiplication with the padded zero values
\cite{krell2022efficient}.

\begin{figure}[t]
  \centering
  \includegraphics[width=\linewidth, page=1]{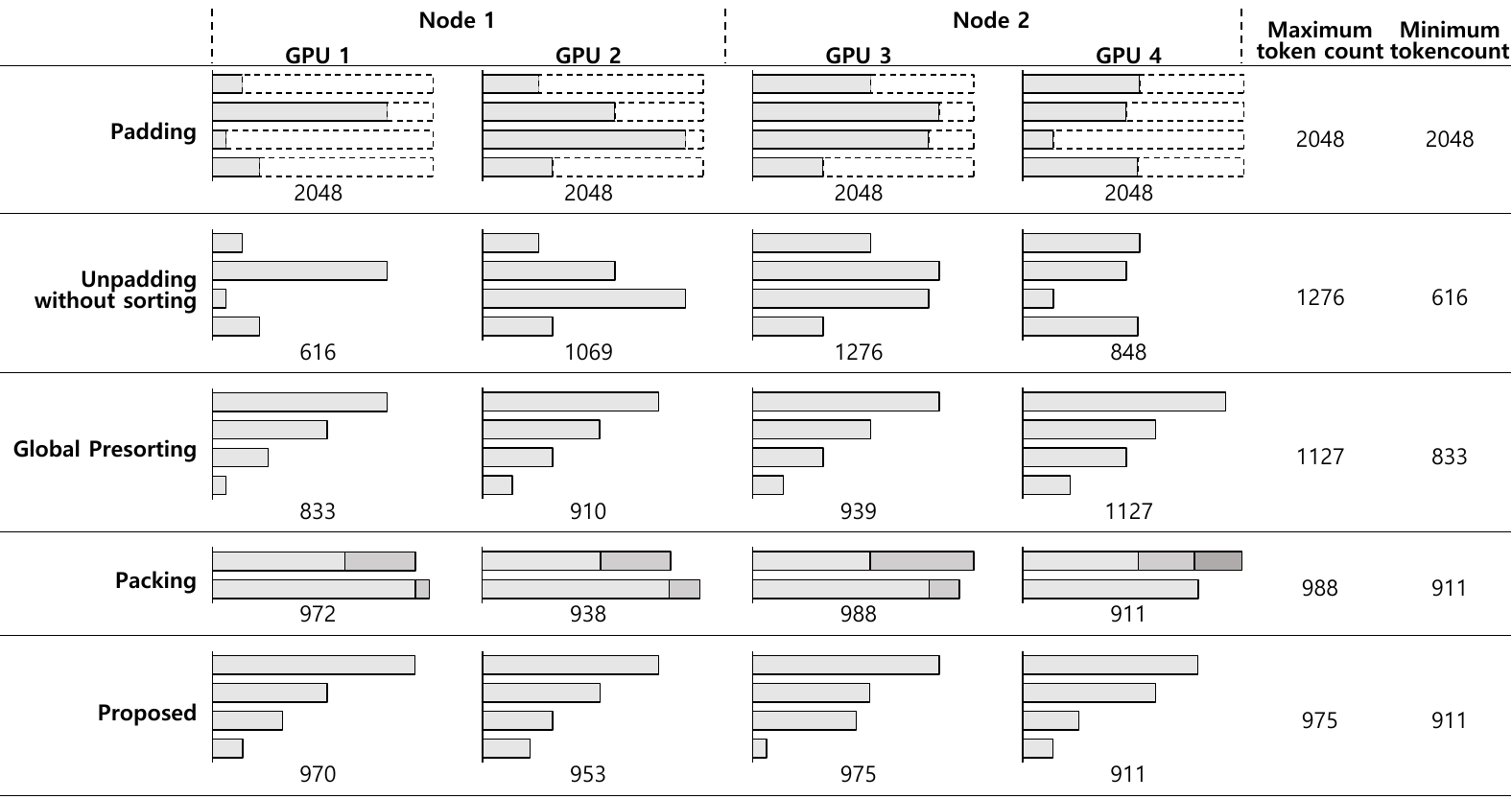}
  \caption{Illustrative comparison of load balancing. Each horizontal bar represents a sequence and its length is proportional to the number
of tokens of the sequence. Numbers below bars represent per-GPU token counts, i.e., the total sum of sample token counts processed by
the associated GPU. Larger difference between maximum and minimum token count implies higher imbalance.}
  \label{fig:motiv_lb_example}
\end{figure}

Figure~\ref{fig:motiv_lb_example} illustrates the wasted computation due to zero padding,
and three existing methods to address this problem. In the figure, the
padding case does not have load imbalance problem, but it
suffers from low utilization of compute resource due to the
wasted computation denoted with dashed rectangles. The
unpadding case illustrates the effect of recently available
unpadding method which performs only the necessary computation
with the given input sequence~\cite{zhai2022bytetransformer}.

That is, it skips the computation of zero padding input in the
padding case. In such a case, as the figure shows, the total
amount of computation can be significantly reduced. However,
the load imbalance across GPUs (the ratio of maximum
to minimum per-GPU token count) is still large, 2.07.

In the case of unpadding with global presorting~\cite{web:mlperfv11nvidia},
the load imbalance can be mitigated as shown in the
figure. It gathers all the input samples of a batch, sort them
in terms of sample length, and allocate them to GPUs in
the sorted order. Thus, the variation of the per-GPU token
counts can be reduced thereby improving the imbalance in
the example. However, this method incurs an additional cost
of global sorting which involves communication across all
the GPUs.

The figure shows that packing~\cite{krell2022efficient} can further
improve load balance. It is to allow the allocation of
more than one samples to the input sequence (of the BERT
model). As the figure illustrates, packing has a potential of
offering well-balanced sample allocations across GPUs.

The figure also illustrates the effect of our proposed method
which, unlike global presorting, does not incur the cost of
global communication, but instead performs local sorting on
a GPU basis. As exemplified in the figure, it has a potential
of obtaining well-balanced sample allocation across GPUs
based on the dataset stratification.

\subsection{Gradient Clipping After/Before AllReduce}
\label{sec:motiv:gc}
\begin{figure}[t]
    \centering
    \subfigure[Gradient clipping after allreduce]{
        \includegraphics[width=62mm]{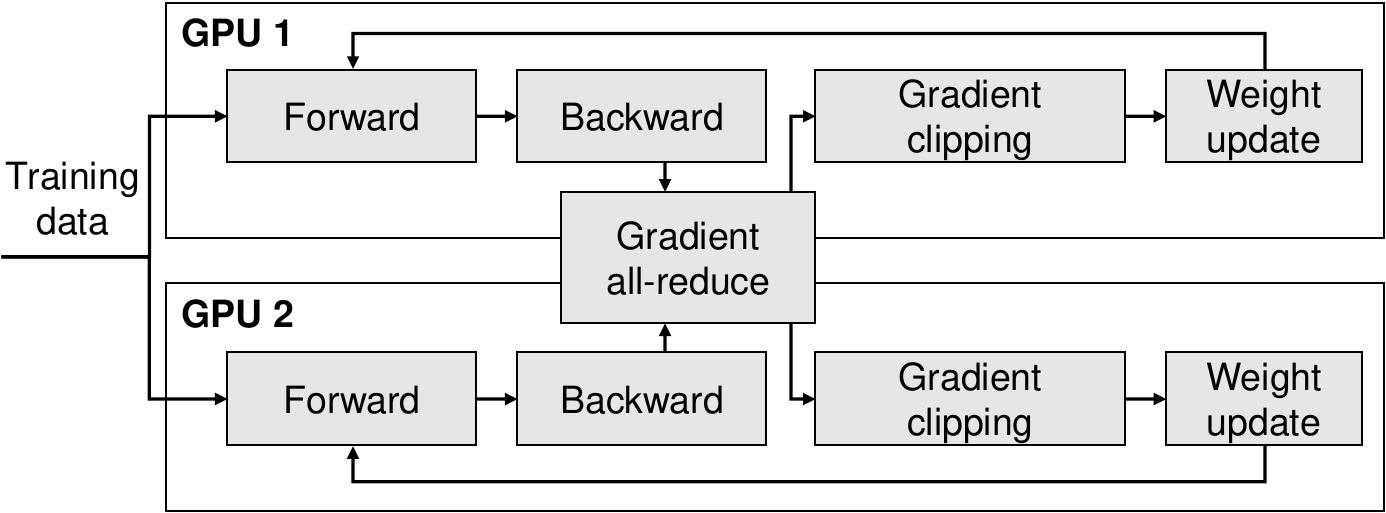}
    }
    \subfigure[Gradient clipping before allreduce]{
        \includegraphics[width=73mm]{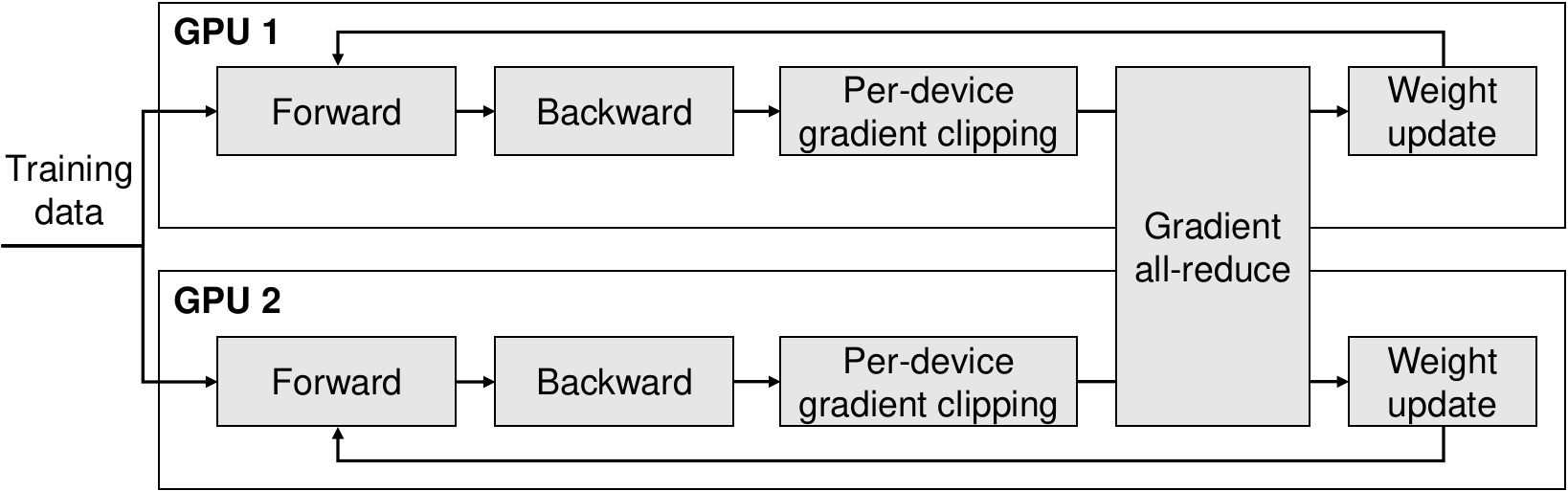}
    }
    \caption{An illustration of two possible ways to implement gradient clipping in distributed training:
    (a) gradient clipping after allreduce and (b) gradient clipping before allreduce.}
    \label{fig:motiv_gc_process}
\end{figure}

Figure~\ref{fig:motiv_gc_process} illustrates two ways of gradient clipping: clipping
after/before allreduce. As the figure shows, gradient
clipping after allreduce allows computation to overlap with
communication since the clipping is applied, on each GPU,
after the final gradients are obtained via communication.
However, in terms of sample efficiency (i.e., the number of
training samples to reach the same level of training quality),
as reported in MLPerf v0.7, gradient clipping before
allreduce tends to outperform gradient
clipping after allreduce. It is mainly because gradient clipping before allreduce can prevent some
problematic mini-batch from having too large impact on
the updated parameters. \DEL{In our experiments, we also confirm
the advantage of gradient clipping before allreduce,
especially with the large batch sizes, e.g., 16K. }Unfortunately,
gradient clipping before allreduce does not allow the
overlap of communication and computation. The benefit
of better sample efficiency can be offset by the lack of
communication/computation overlap. Thus, gradient clipping
before allreduce is rarely adopted in large-scale distributed
training. In our work, we propose applying both gradient
clipping before allreduce and the overlap of computation
and communication by performing gradient clipping at the
granularity of bucket.

\section{Proposed Method}
\label{sec:opt}

\subsection{Local Presorting with Dataset Stratification}
\begin{figure}[b]
  \centering
  \includegraphics[width=3in, page=1]{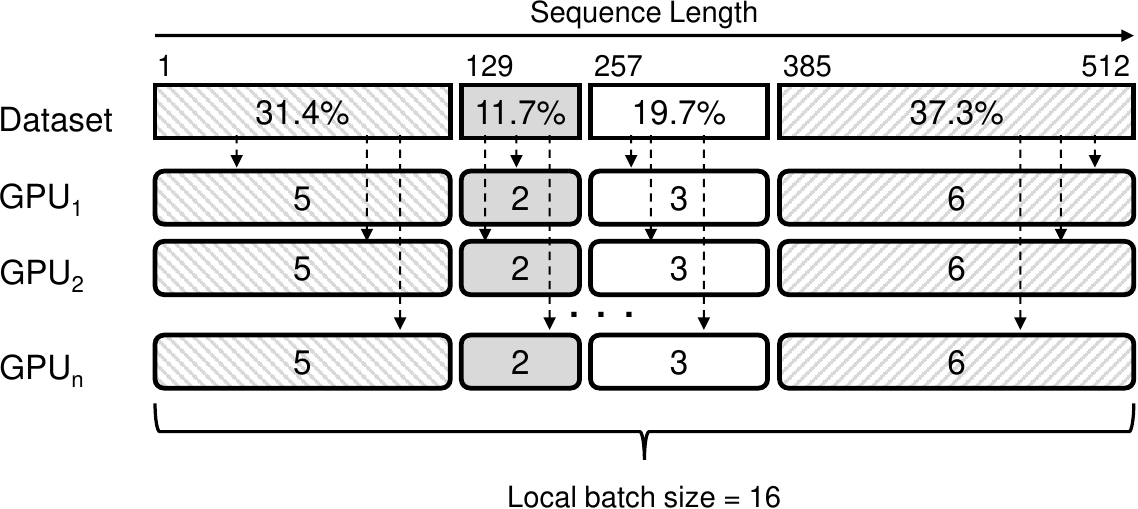}
  \caption{Example of stratification when dataset is stratified into
four strata and batch size is 16.}
  \label{fig:opt_stratification}
\end{figure}

Stratification is a well-known variance reduction technique
which assigns samples to groups called strata (groups
formed in terms of sequence length in our case) and selects
samples in proportion to the probabilities of strata. Figure~\ref{fig:opt_stratification}
exemplifies our proposed stratification applied to Wikipedia
dataset. We assume batch size of 16 and maximum sequence
length of 512. First, we use four strata in terms of sequence
length, i.e., stratum 1 contains samples of length from 1 to
128, stratum 2 from 129 to 256, stratum 3 from 257 to 384
and stratum 4 from 385 to 512. As the figure shows, each
stratum has its own probability determined by the number
of samples belonging to it. For instance, the larger probabilities
of strata 1 and 4 exhibit the bi-modal characteristics,
i.e., mostly short or long sequences, of Wikipedia dataset.
As the figure shows, in our example of local batch size of
16, we can select 5, 2, 3 and 6 samples on strata 1, 2, 3, and
4, respectively. Note that, given the local batch size, the
number of selected samples per stratum is determined by
the probability of the stratum. As the example shows, the
sampling with stratified dataset can ease the load imbalance
problem.

\begin{figure}[t]
  \centering
  \includegraphics[width=3.5in, page=2]{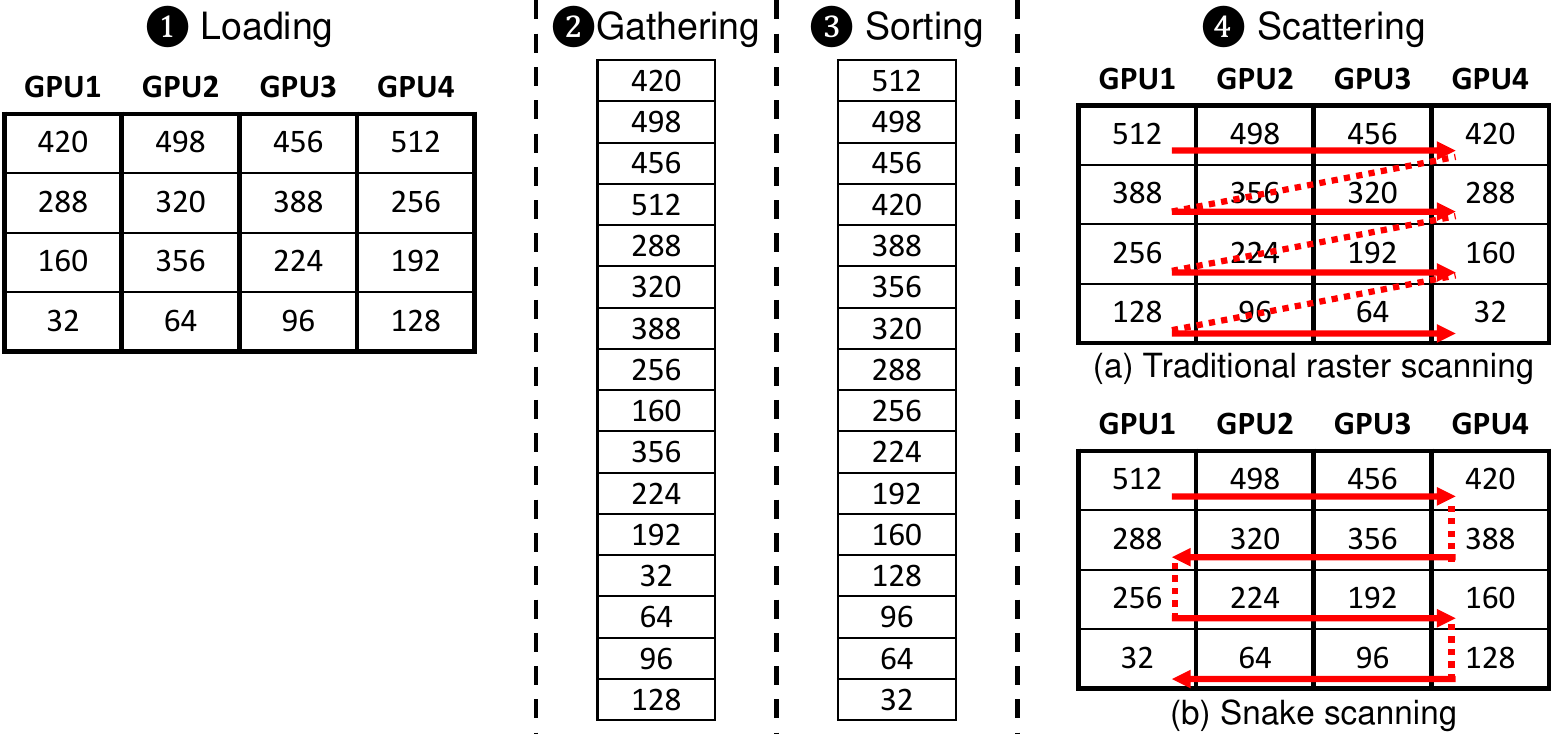}
  \caption{The example of local presorting with two different retrieving
patterns: (a) raster and (b) snake scanning.}
  \label{fig:opt_presorting}
\end{figure}

We propose applying stratification in a GPU node basis.
In the existing method of global presorting~\cite{web:mlperfv11nvidia},
presorting across all GPU nodes incurs high communication
overhead since it gathers all samples from all the GPU nodes
via slow inter-node communication. Our proposed per-GPU
node stratification enables us to presort samples only inside
of GPU node, which we call \textit{local presorting}, thereby
avoiding expensive inter-node communication.

Figure~\ref{fig:opt_presorting} exemplifies how our local presorting works. We
assume a GPU node consists of 4 GPUs and each GPU is initially
assigned four samples (four values correspond to their
lengths). In the steps of gathering and sorting, we all-gather
all the samples inside of a GPU node and sort them. When
scattering the presorted samples, we also propose applying
a snake pattern scanning as shown in the figure. Compared
with the conventional raster scanning, our experiments show
the snake pattern proves more effective in mitigating load
imbalance.

\subsection{Bucket-wise Gradient Clipping before AllReduce}
\begin{figure}[b]
  \centering
  \includegraphics[width=5.2in]{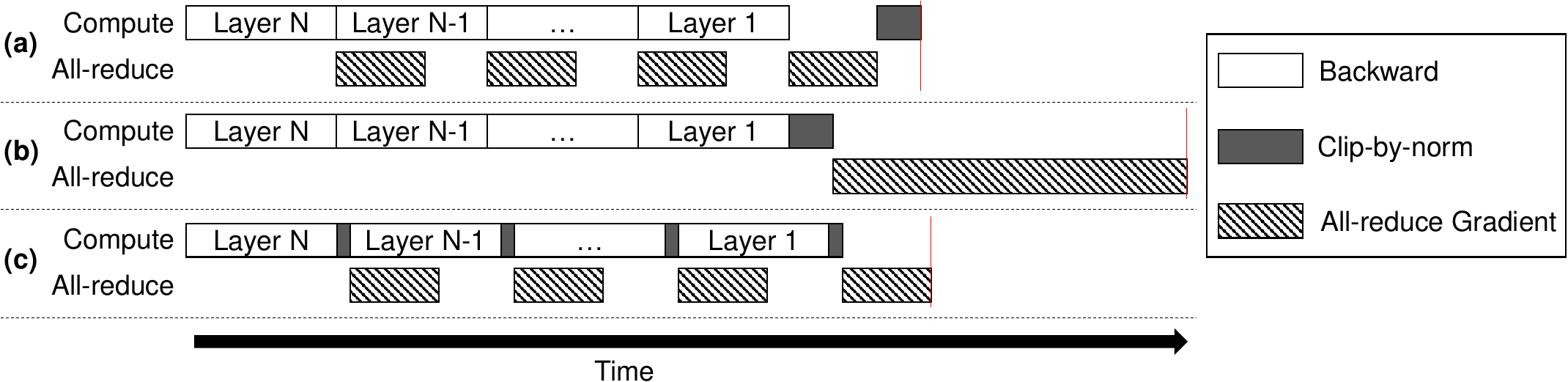}
  \caption{Timeline of gradient clipping methods: (a) gradient clipping
after allreduce, (b) gradient clipping before allreduce, and (c)
bucket-wise gradient clipping.}
  \label{fig:motiv_gc_timeline}
\end{figure}

Figure~\ref{fig:motiv_gc_timeline} (a) and (b) compare how gradient clipping can be
applied (a) after and (b) before allreduce under bucket-based
synchronization. Gradient clipping after
allreduce still enables the overlap of computation and communication.
However, gradient clipping before allreduce
does not benefit from the overlap. Figure~\ref{fig:motiv_gc_timeline} (c) shows how
our proposed bucket-wise gradient clipping before allreduce
enables the overlap of communication and computation. As
shown in the figure, gradients in each bucket are clipped as
soon as they are ready and then immediately synchronized
with allreduce. Since the clipping is independently applied
to each bucket on each GPU, we can transfer buckets of
clipped gradients during back-propagation.
The pseudocode is provided in the Appendix.

\section{Experiments}
\label{sec:eval}

\textbf{Experimental environment~~~} We train the BERT model
on a large-scale GPU training system consisting of 1,024
NVIDIA A100 GPUs with 100 GB/s network bandwidth.
More precisely, the system is a cluster of 128 nodes each of
which consists of two AMD EPYC 7543 CPUs and eight
NVIDIA 80GB A100 GPUs connected by NVLink and
NVSwitch. The servers in the cluster are connected by four
HDR 25GB/s Infinibands. We use NVIDIA PyTorch
container image, release 21.06 (21.09) for MLPerf v1.1 and
(v2.0).

We use automatic hyperparameter optimization tool,
Neural Network Intelligence~\cite{microsoftnni}, to compare optimizers
and gradient clipping methods in a fair way. We
use SMAC~\cite{hutter2011sequential} and reduce the search space
by selecting important parameters and exploiting locality in
search space as explained in Appendix.

\subsection{MLPerf BERT Benchmark Results}
Table~\ref{tab:mlperf11} shows the large-scale training performance of BERT
reported in MLPerf Benchmark. Ours ranks the first in the
configuration of 1,024 accelerators both in v1.1 and v2.0.

\begin{table}\footnotesize
  \centering
  \caption{Published results from MLPerf Training
  v1.0~\cite{arxiv:2019:mlperf}, v1.1~\cite{web:mlperfv11} and  v2.0~\cite{web:mlperfv20}}

  \begin{tabular}{c c c c c | c c}
      \toprule
      \textbf{Version}
      & \textbf{Vendor}
      & \textbf{Accelerator}
      & \textbf{\#}
      & \textbf{Batch size}
      & \textbf{Results (seconds)}
      & \textbf{\# of samples}
      \\
      \midrule
      v1.0 & NVIDIA    & NVIDIA A100 GPU     & 1024 & 3072  & \hspace{0.9em}43.5   & 2.6M$\sim$3.0M \\ %
           & NVIDIA    & NVIDIA A100 GPU     & 4096 & 12288 & \hspace{0.9em}19.0   & 4.7M$\sim$5.0M \\ %
           & Google    & TPU-v4              & 2048 & 6144  & \hspace{0.9em}19.1   & 3.2M$\sim$3.4M \\ %
           & Google    & TPU-v4              & 3456 & 6912  & \hspace{0.9em}16.5   & 3.4M$\sim$3.6M \\ %
      \midrule
      v1.1 & NVIDIA    & NVIDIA A100 GPU     & 1024 & 3072  & \hspace{0.9em}33.5   & 2.6M$\sim$3.0M \\ 
           & NVIDIA    & NVIDIA A100 GPU     & 4320 & 12960 & \hspace{0.9em}13.6   & 4.4M$\sim$5.0M \\ 
           & Microsoft & NVIDIA A100 GPU     & 1024 & 3072  & \hspace{0.9em}39.4   & 2.6M$\sim$2.8M \\ 
           & Microsoft & NVIDIA A100 GPU     & 2048 & 6144  & \hspace{0.9em}25.3   & 3.2M$\sim$3.4M \\ 
           & \textbf{Ours}   & \textbf{NVIDIA A100 GPU}    &
           \textbf{1024} & \textbf{16384} & \hspace{0.95em}\textbf{25.1}
           & \textbf{2.9M$\sim$3.3M}  \\ 
      \midrule
      v2.0 & NVIDIA    & NVIDIA A100 GPU     & 1024 & 4096  & \hspace{0.9em}25.3   & 2.6M$\sim$3.0M \\
           & NVIDIA    & NVIDIA A100 GPU     & 4096 & 16384 & \hspace{0.9em}12.4   & 5.2M$\sim$5.9M \\
           & Google    & TPU-v4              & 3456 & 6912  & \hspace{0.9em}13.7   & 4.8M$\sim$4.9M \\
           & Google    & TPU-v4              & 4096 & 14336 & \hspace{0.9em}11.0   & 6.5M$\sim$6.6M \\
           & \textbf{Ours} & \textbf{NVIDIA A100 GPU} &
           \textbf{1024} & \textbf{16384} & \hspace{0.9em}\textbf{22.3}   &
           \textbf{2.9M$\sim$3.0M} \\
      \bottomrule
  \end{tabular}
  \label{tab:mlperf11}
\end{table}

NVIDIA continuously improves the unpadding Fused Multi
HEAD Attention (FMHA) kernel~\cite{web:apex} and uses
different optimization strategy in each benchmark. For
load balancing, NVIDIA uses unpadding FMHA kernel
and global presorting for small-scale systems ($\le$64
GPUs). They use pad FMHA for large-scale system ($\ge$1,024 GPUs)
until v1.1 and change to unpad FMHA and
packing in v2.0. The gradient clipping method is changed
from before allreduce to after allreduce from v1.1.

Our internal optimization history is summarized in Table~\ref{tab:eval_history}.
When we reproduced NVIDIA v1.0 in our system, there
was significant performance gap, 43.5 vs 59.5 seconds. The
network bandwidth difference (200 vs 100 GB/s) and lack
of SHARP (scalable hierarchical aggregation and reduction protocol)
affected the performance because the batch
size per GPU was very small at 3 and communication and
computation was not overlapped due to gradient clipping
before allreduce. To overcome this network bandwidth
limitation, we first changed the gradient clipping method
from before allreduce to after allreduce and tried to increase
batch size as much as possible for high GPU utilization.
After extensive hyperparameter optimization, we found that
conventional ADAM works better than LAMB for large
batch training. Then we applied proposed load balancing
and gradient clipping method. We used the same method for
v2.0 except for updating SW stack such as CUDA, NCCL,
and APEX.

\begin{table}\footnotesize
  \centering
  \caption{Summary of performance optimization history on 1024 NVIDIA
  A100 GPUs for MLPerf v1.1}
  \label{tab:eval_history}
  \begin{tabular}{l | r r r}
    \toprule
      \textbf{Optimization Item} & \textbf{Time (sec)} & \textbf{Batch size} &
      \textbf{\# of samples}\\
    \midrule
      NVIDIA v1.0 reported & 43.5 & 3K                  & 2.6$\sim$3.0M\\
      NVIDIA v1.0 reproduced & 59.5 & 3K                & 2.6$\sim$3.0M \\
      + PyTorch DDP + ADAM & 34.5 & 16K                 & 3.9$\sim$4.3M \\
      + Bucket-wise gradient clipping & 28.9 & 16K      & 2.9$\sim$3.3M \\
      + Local Presorting + Stratification & 25.1 & 16K  & 2.9$\sim$3.3M \\
    \bottomrule
  \end{tabular}
\end{table}

\subsection{Irregular Sequence Length Handling}
In this subsection, we evaluate methods for handling irregular
sequence lengths. Note that global presorting, packing,
and our proposed method are based on unpadding.

\textbf{Load balancing effect simulation~~~} First, the effectiveness
of load balancing is evaluated by simulation. We basically
assume that effective batch size per GPU (local batch
size) is 16. We also assume packing ratio 2 for packing,
and 8 GPUs per node for our proposed method. In
packing/our/other methods, we randomly retrieves 8/16/16 sequences
for each GPU from pre-packed/stratified/original
dataset, respectively. After retrieving data, we balance them
following each load balancing method and find maximum
and minimum token count processed on a GPU. We repeat
the above simulation 100,000 times and calculate average
maximum and minimum token count.

\begin{figure}
  \centering
  \includegraphics[width=4.2in]{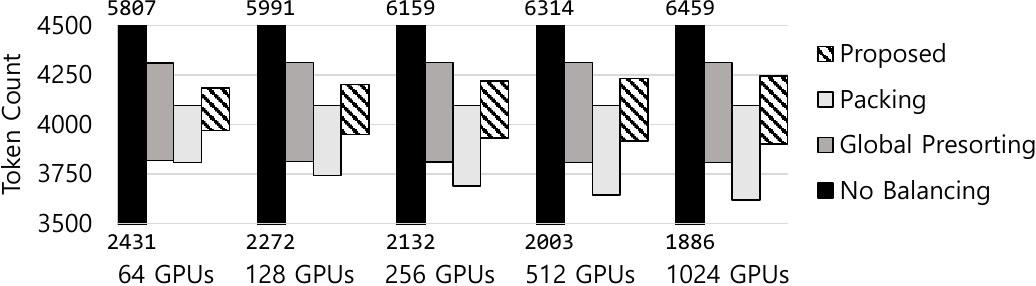}
  \caption{Average range of token count per GPU for each load
balancing method when effective local batch size is 16. The bottom
(top) of bar represents the average minimum (average maximum)
token count when using the associated load balancing method. The
maximum token count of a sequence is 512.}
  \label{fig:eval_lb_range}
\end{figure}

Figure~\ref{fig:eval_lb_range} depicts the average token count range of each
load balancing method, where the shorter bar implies the
better balancing. The method of no balancing shows a critical
imbalance problem even in 64 GPUs, and the packing
method can also suffer from the imbalance problem as the
system size grows. In addition, the packing method can also
suffer from low utilization due to limited maximum token
count.
Please refer Appendix for load balancing improvement according to stratification/local presorting/snake scanning step by step.

\begin{figure}
    \centering
    \subfigure[Latency with local batch size 6 and 12.]{
        \includegraphics[width=65mm]{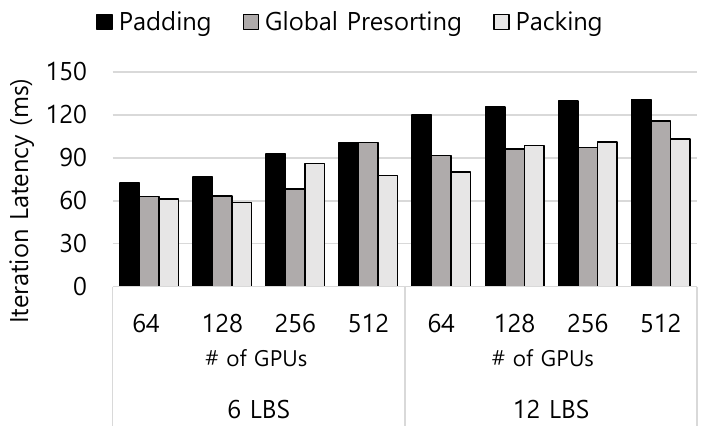}
        \label{fig:eval_lb_latency_small_batch}
    }
    \subfigure[Latency with local batch size 16 and 32.]{
        \includegraphics[width=65mm]{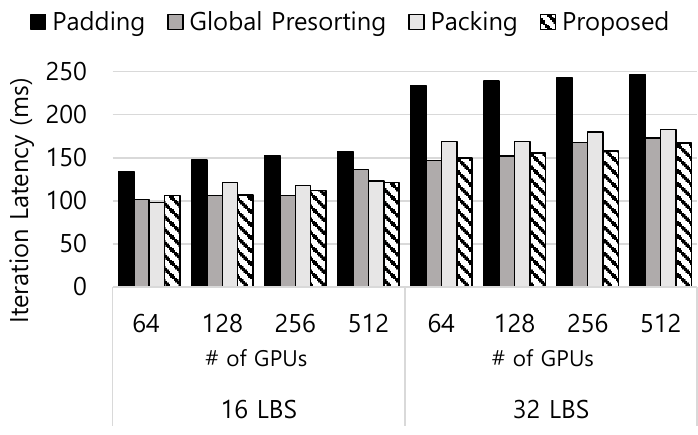}
        \label{fig:eval_lb_latency_large_batch}
    }
    \caption{Latency of each load balancing method with small and large local batch size.}
\end{figure}

\textbf{End-to-end latency comparison~~~} With padding as the
baseline, global presorting, packing and our method are
compared in terms of end-to-end latency. Figure~\ref{fig:eval_lb_latency_small_batch} shows
the latency of each method according to the effective local
batch size and the number of GPUs. Packing shows good
performance when effective local batch size is small (6 and
12) with the help of CUDA graph. The use of CUDA graph
by packing is enabled because packing always launches the
same FMHA kernel, which is determined by maximum sequence
length in the local mini batch. On the other hand,
global presorting cannot use CUDA graph acceleration because
it needs to launch various FMHA kernels according to
the length of sequences to leverage the advantage of the
unpadding. Note that we do not evaluate the proposed
method with small local batch size because stratification
cannot reflect the ratio of strata for small local batch size.

Figure~\ref{fig:eval_lb_latency_large_batch} shows the training
latency of load balancing methods
with large batches. When local batch size gets larger
than 12, packing could not use CUDA graph like other method because of out-of-memory.
Therefore, packing loses advantage over global presorting
and global presorting runs faster than packing for
most cases. The proposed method gives the lowest latency
when the number of GPUs gets larger, because it eliminates communication over
all GPUs for load balancing. This result implies that our
proposed method is suitable for large batch and large-scale GPU
environments.

We also profile the latency for load balancing and calculate its
proportion in the total training latency. In case of 1,024 GPUs,
the global presorting takes 18.2\% of the total training time,
while our method takes only 1.9\%. Refer to Appendix for detailed analysis of this experiment.

\vspace{4mm}
\subsection{Gradient Clipping}
\label{sec:eval:gc}
\textbf{Sample efficiency~~~} In this subsection, we compare three gradient
clippings: gradient-clipping-after-allreduce, gradient-clipping-before-allreduce, and bucket-wise gradient clipping.
First, we compare the sample efficiency by measuring
the accuracy for each gradient clipping method after training
2.9 million samples. Because the optimal hyperparameters
could be different between gradient clipping methods, the
hyperparameters for each method are searched through automated
hyperparameter optimization for a fair comparison.
Refer to Appendix for the automated hyperparameter optimization
methodology for this experiment.

In Table~\ref{tab:eval_gc_accuracy}, the higher accuracy implies the better sample efficiency.
As we discussed in Section~\ref{sec:motiv:gc}, gradient-clipping-after-allreduce
shows lower sample efficiency due to the
negative effects of problematic mini-batch having large gradients.
On the other hand, bucket-wise gradient clipping
method shows high sample efficiency as much as
gradient-clipping-before-allreduce, because it reduces the impact
of problematic mini-batch effectively as well as
gradient-clipping-before-allreduce.

\begin{figure}
  \centering
  \includegraphics[width=3.4in]{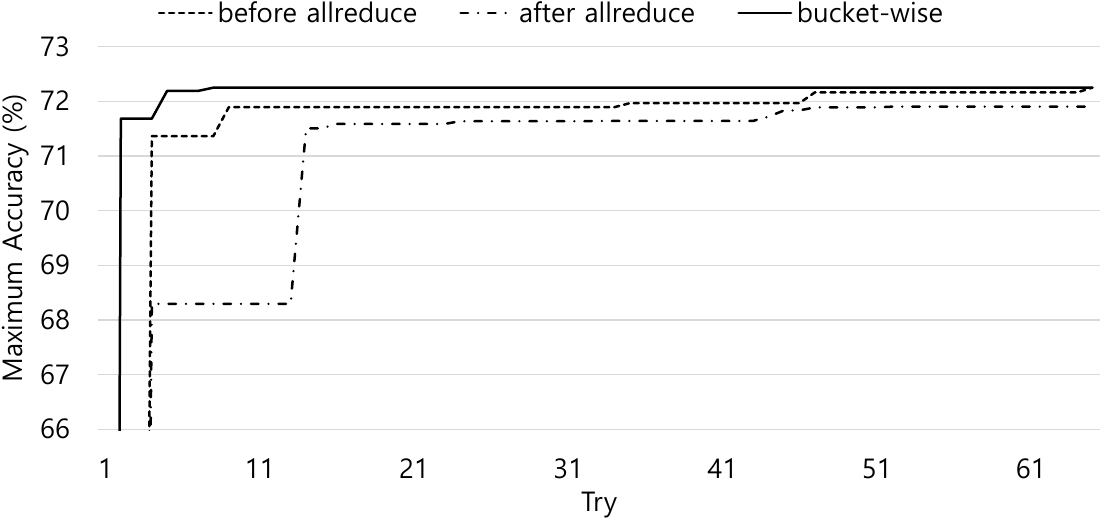}
  \caption{Comparison of hyperparameter optimization cost of gradient clipping methods.}
  \label{fig:eval_gc_hpo_cost}
\end{figure}

\textbf{Hyperparameter optimization cost~~~} In order to compare the cost of finding the optimal hyper-parameter for each gradient clipping method, we measure the maximum accuracy according to the number of hyperparameter optimization attempts.
For fair comparison, we used Microsoft Neural Network Intelligence (NNI) and SMAC for automatic hyperparamter optimization.
In the Figure~\ref{fig:eval_gc_hpo_cost}, y-axis means the maximum accuracy achieved over each given number of tries.
The figure depicts that bucket-wise gradient clipping requires far fewer attempts to obtain hyperparameters for reasonable accuracy.
To achieve the accuracy in Table~\ref{tab:eval_gc_accuracy}, we have run 9, 66, and 130 tries for bucket-wise gradient-clipping, gradient-clipping-before-allreduce, and gradient-clipping-after-allreduce, respectively.

\begin{table}\footnotesize
  \centering
  \caption{MLM accuracy when using each gradient clipping method.
Global batch size = 16384, Local batch size = 16, The number of samples = 2.9M}
  \label{tab:eval_gc_accuracy}
  \begin{tabular}{l | r}
    \toprule
      \textbf{Gradient Clipping Method} & \textbf{Accuracy (\%)} \\
    \midrule
      \textbf{Gradient Clipping After Allreduce}  & 72.05 \\
      \textbf{Gradient Clipping Before Allreduce} & 72.26 \\
      \textbf{Bucket-wise Gradient Clipping}      & 72.25 \\
    \bottomrule
  \end{tabular}
\end{table}

\begin{figure}
  \centering
  \includegraphics[width=4.9in]{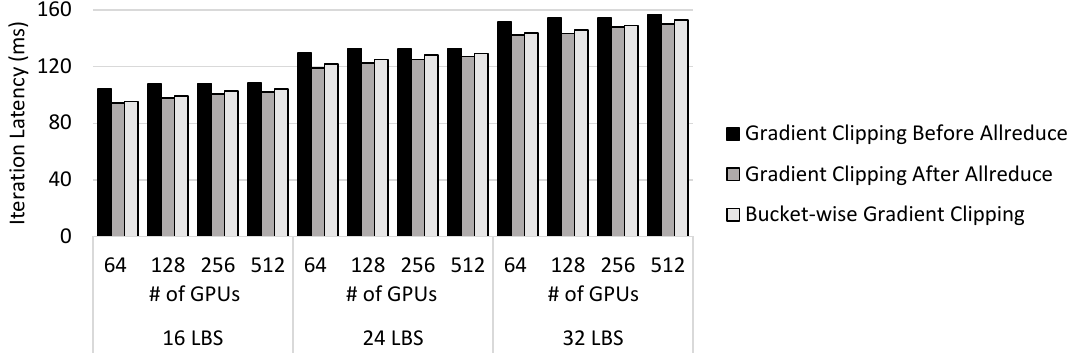}
  \caption{Latency comparison between gradient clipping methods.}
  \label{fig:eval_gc_latency}
\end{figure}

\textbf{Overhead~~~} We also compare the throughput degradation
due to gradient clipping methods.
We measure the latency of single training iteration under
gradient clipping. In Figure~\ref{fig:eval_gc_latency}, we can observe that
bucket-wise gradient clipping and gradient-clipping-after-allreduce
shows similar latency since both methods well
hide the overhead of gradient allreduce. On the other hand,
gradient-clipping-before-allreduce shows higher latency because
it cannot hide gradient allreduce overhead.

\section{Discussion and Limitations}
\textbf{Other parallelisms across transformer based models~~~}
The proposed data-parallel training method can be applied together with other parallelisms.
For instance, in case of DeepSpeed where model, pipeline and (ZeRO) data
parallelisms are utilized, our gradient clipping method can be applied to improve the sample
efficiency of data parallelism thereby contributing to fast training.

\textbf{Comparison with previous distributed gradient clipping techniques~~~}
M. Liu \textit{et al.}~\cite{liu2022communication}consider a local SGD-type method;
since it allows multiple steps of each worker to run before communicating with the others, the local gradient clipping is a natural choice when adopting gradient clipping.
If the method aggregates model paparameters on all machines after every single step of local updates, this is identical to the gradient-clipping-before-allreduce, which is compared with the proposed bucket-wise gradient-clipping in our paper.
In addition, W. Wen \textit{et al.}~\cite{wen2017terngrad} and Y. Lin \textit{et al.}~\cite{lin2017deep} proposed other gradient-clipping approaches, which can also be categorized in the gradient-clipping-before-allreduce type ones; on the other hand, to the best of our knowledge, our bucket-wise gradient-clipping is the first to combine the advantages of the gradient-clipping-after-allreduce and the gradient-clipping-before-allreduce and take the advantages of both.
By doing this, the bucket-wise gradient-clipping allows to overlap communication and computation while achieving an algorithmic efficiency of the gradient-clipping-before-allreduce level.

\textbf{Small batch size~~~}
Our load balancing method may have a limited utility in case of very small
batches (e.g., local batch size of 4) since stratification is based on the
probabilities of bins (i.e., strata) and, in case of a very small
(e.g., a single) number of samples per bin. In such a case, there is a risk that sampling cannot always reflect the bin statistics. We found that our load
balancing becomes effective when the local batch size is 16 where the
numbers of per-stratum samples are 5:2:3:6 on four strata, which proves to
faithfully reflect the statistics of our training data.
On the other hand, bucket-wise gradient clipping is applicable to small batch training since its working unit is a bucket and the number of buckets depends on the model size, not the batch size. We conducted a similar experiment as Table~\ref{tab:eval_gc_accuracy} but with a small batch size (256) and hyperparameter from the BERT paper. The proposed method shows higher accuracy (71.76\%) than gradient clipping after allreduce (71.56\%) after training with the same number of samples.

\section{Conclusion}
We addressed two key issues in large-scale distributed
training of BERT models, load balancing and
computation/communication overlap. In order to account for the
diverse lengths of sentences in the training dataset and to
reduce pre-sorting cost, we proposed local presorting based
on dataset stratification. We also proposed bucket-wise gradient
clipping before allreduce, which enables us to benefit
from both gradient clipping before allreduce and the overlap
of gradient computation and synchronization. In our cluster
of 1,024 NVIDIA A100 GPUs, we report the state-of-the-art
training time of 25.1 (22.3) seconds to finish MLPerf BERT
training, which is 1.33$\times$ (1.13$\times$) and 1.57$\times$ faster than the
other top two (one) submissions to MLPerf v1.1 (v2.0).


\newpage
\appendix

%
%
\section{Automatic Hyperparameter Optimization}
\label{appendix:hpo}
\citeSupp{choi2020empirical} and \citeSupp{schmidt2021descending} explained
challenges in fairly comparing optimizers and revealed that
hyperparameter tuning protocol is a key determinant of
optimizer rankings. Currently, various hyperparmater optimization
algorithms such as SMAC \citeSupp{hutter2011sequential_supp},
TPE \citeSupp{bergstra2015hyperopt}, BOHB \citeSupp{falkner2018bohb},
and GP Tuner \citeSupp{snoek2012practical} are available in tools such
as Hyperopt \citeSupp{bergstra2015hyperopt}, Spearmint \citeSupp{snoek2012practical},
Autotune \citeSupp{koch2018autotune}, Optuna \citeSupp{akiba2019optuna},
Vizier \citeSupp{golovin2017google}, and Microsoft Neural
Network Intelligence (NNI) \citeSupp{microsoftnni_supp}.

However, there is no agreement on how to conduct an experiment
that fairly compares learning algorithms. Although it
may not be perfect, we perform hyperparmater optimization
with the same computation budget when comparing optimizers
(ADAM vs LAMB) and gradient clipping methods
mentioned in Section~\ref{sec:eval:gc}. Among various tools and hyperparameter
optimization algorithms, we simply select Microsoft
NNI and SMAC because of its popularity, and then try to
use computation budget for hyperparmater optimization efficiently
by following two approaches:

\paragraph{Importance-aware search target selection} We measure
the importance of hyperparameters by fANOVA \citeSupp{hutter2014efficient}
values and observe that some hyperparameters
such as the ratio between peak and end learning rate (LR), $\epsilon$,
and LR init\footnote{LR init is a coefficient that determines the initial learning rate.}
have little impact on the training performance.
Hence, it is more beneficial to exclude those less-important
hyperparameters in the search procedure because of limited
time budgets and computation costs. The exclusion of
less-important hyperparameters helps us to decrease the dimension
of the search space. Table~\ref{tab:eval_hpo_range} exemplifies the result
of hyperparameter optimization.

\begin{figure}[h]
  \centering
  \includegraphics[width=4in]{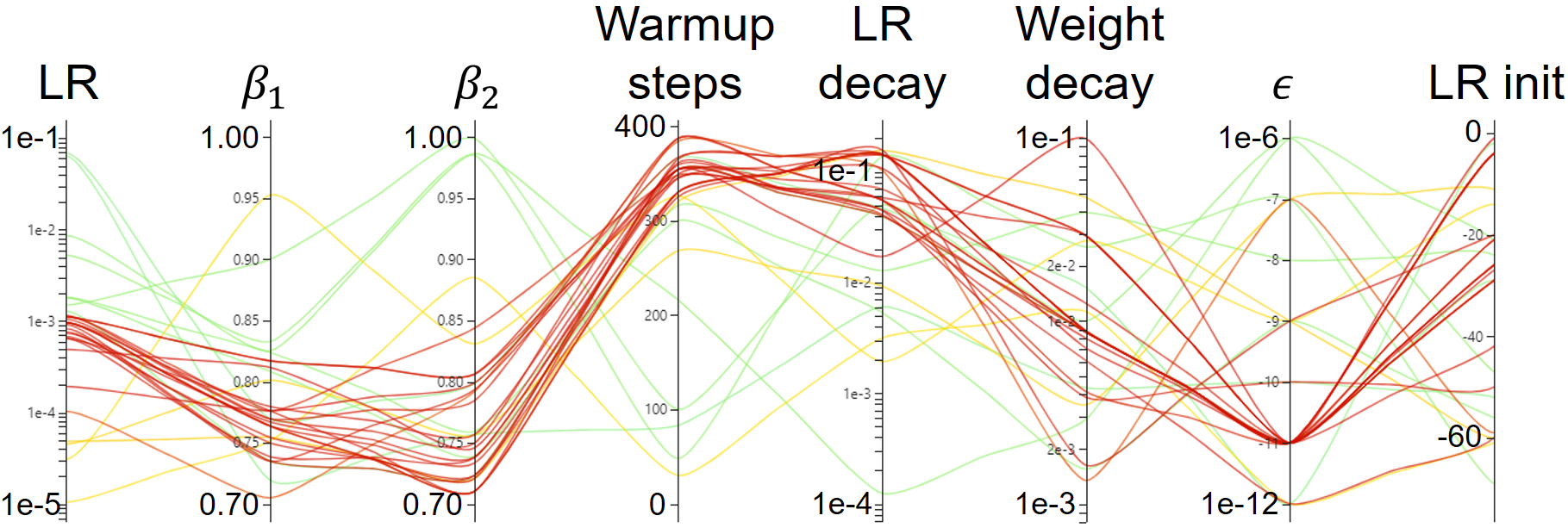}
  \caption{Illustration of hyperparameter searching results.}
  \label{fig:motiv_hpo_result}
\end{figure}

\paragraph{Search range reduction of the selected targets} In addition
to the selection of search targets, it is also important to
properly choose the search ranges of the selected hyperparameters.
If the search ranges are too wide, it takes too much
time to search optimal hyperparameters. On the other hand,
if the search ranges are too narrow, the HPO tools may find
suboptimal hyperparameters. In Figure~\ref{fig:motiv_hpo_result}, each line stands
for combinations of the hyperparameters evaluated by the
NNI. The green lines indicate the combinations that show
low task performance, whereas the red bold lines indicate
the good candidates for the task. As observed in the figure,
there exists a cluster that shows high task performance and
such a locality allows us to narrow down the search space
without the risk of losing optimal points.

\begin{table}[h]
\footnotesize
  \centering
  \caption{Example of hyperparameter search space reduction. We
set hyperparameter optimization budget as 150 and 50 trials for
coarse and fine stages, respectively}
  \label{tab:eval_hpo_range}
  \begin{tabular}{l | c | l | l}
    \toprule
      \textbf{Hyperparameter} & \textbf{fANOVA} & \textbf{Coarse} & \textbf{Fine} \\
    \midrule
      LR                      & 0.50            & [1e-5, 1e-1]    & [1e-3, 1e-3]  \\
      $\beta_{1}$             & 0.25            & [0.7, 0.999]    & [0.7, 0.80]   \\
      $Weight decay$          & 0.17            & [1e-3, 1e-1]    & [1e-2, 1e-1]  \\
      $\beta_{2}$             & 0.03            & [0.7, 0.999]    & [0.85, 0.999] \\
      Warmup steps            & 0.03            & [0, 400]        & [0, 300]      \\
    \midrule
      End LR ratio            & 0.01            & [1e-4, 0.2]     & 5e-4          \\
      $\epsilon$              & 0.01            & [1e-12, 1e-5]   & 1e-11         \\
      LR init                 & 0.00            & [-73, 0]        & -47           \\
    \bottomrule
  \end{tabular}
\end{table}

%
%

\section{Bucket-wise Gradient Clipping Algorithm}

Algorithm~\ref{alg:bucket_wise_gc}
shows the bucket-wise gradient clipping in detail. Note that
the threshold $c$ is divided by $B$ to ensure that bucket-wise
clipping behaves similarly to whole gradient clipping. We
observe training becomes unstable without this.

\begin{algorithm}[h]
    \caption{Bucket-wise Gradient Clipping}
    \begin{algorithmic}[1]
    \STATE {\bfseries Input:} model parameter $\theta$, training data \textit{D}, minibatch sample \textit{X}, number of iteration \textit{T}, gradient \textit{$G$}, bucket of gradients \textit{$G_b$}, number of buckets \textit{$B$}, threshold for clip \textit{$c$}, learning rate $\eta$
      \STATE $\theta \gets$ random initialization
      \FOR {$t=1,...,T$}
        \STATE Sample minibatch $\textit{X} \subset \textit{D}$
        \FOR {$b=B,B-1,\cdots,1$}
          \STATE Compute $G_b$
          \IF {$||G_b|| \geq c/\sqrt{B}$} 
            \STATE $G_b \gets$  $c\frac{G_b}{\sqrt{B}||G_b||}$
            \hspace{5.5em}\raisebox{.5\baselineskip}[0pt][0pt]{$\left.\rule{0pt}{3.2\baselineskip}\right\}\ \mbox{Do in parallel}$}
          \ENDIF
          \STATE $G_b \gets$  All-Reduce($G_b$) 
        \ENDFOR
        \STATE $G \gets [G_1, \cdots, G_B]$
        \STATE $\theta \gets ADAM(\theta, \eta, G)$ 
      \ENDFOR
    \end{algorithmic}
    \label{alg:bucket_wise_gc}
\end{algorithm}

%
%

\section{Extensive Experiments for Load Balancing}
\subsection{Latency}
\begin{figure}[h]
  \centering
  \includegraphics[width=3.1in]{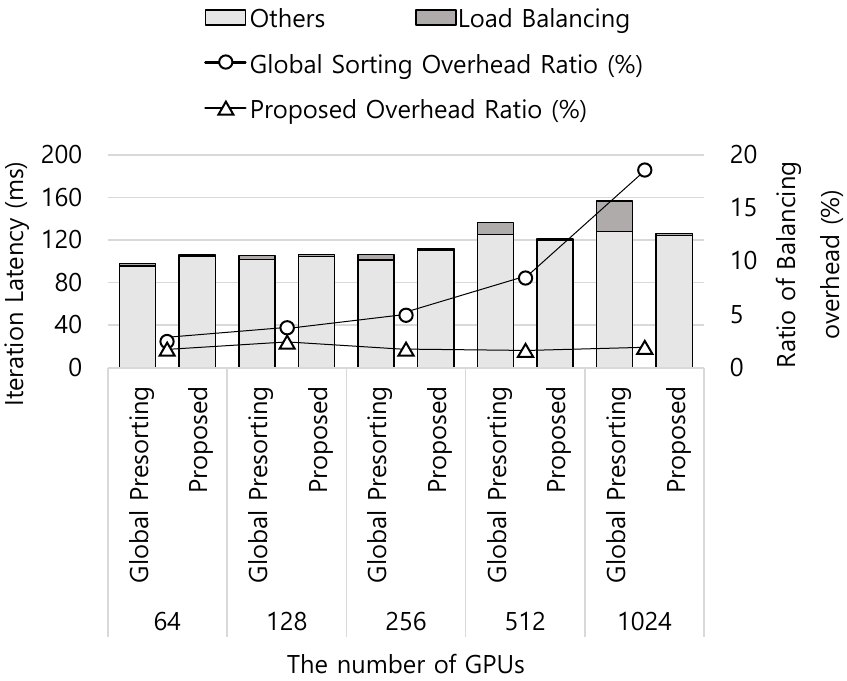}
  \caption{Latency breakdown of global presorting and proposed
method. The latency of load balancing is profiled using
torch.profiler}
  \label{fig:eval_lb_latency_breakdown}
\end{figure}

We profile the latency for load balancing and calculate
its proportion in the total training latency.
We conduct this experiment only for global presorting and the proposed
method, because only these two methods have a stage for
load balancing. On the other hand, packing uses packed
dataset which is prepared before starting training. Figure~\ref{fig:eval_lb_latency_breakdown}
shows that the load balancing overhead of global presorting
becomes larger when more GPUs are used since it requires
all GPUs participate in all-gather and sort the gathered samples.
The load balancing latency takes 18.2 \% of the total
training time in case of 1,024 GPUs. On the other hand, in
our method, the overhead of load balancing remains constant
across different numbers of GPUs since GPUs do not
communicate with each other during load balancing time.

%
%

\subsection{Balance}

Table~\ref{tab:eval_load_balancing} compares global presorting and our proposed methods according to stratification/local presorting/snake scanning step by step. Combining stratification, local presorting,
and snake scanning pattern is required to outperform global
presorting.

\begin{table}[h]
\footnotesize
  \centering
  \caption{Load balancing improvement according to
  stratification/local presorting/snake scanning step by step.}
  \label{tab:eval_load_balancing}
  \begin{tabular}{l | r r}
    \toprule
                              & \textbf{Minimum} & \textbf{Maximum} \\
    \midrule
      No Balancing             & 1,886            & 6,459 \\
      + Stratification         & 3,620            & 4,482 \\
      + Local Presorting       & 3,743            & 4,401 \\
      + Snake Scanning Pattern & 3,900            & 4,246 \\
    \midrule
      Global Presorting        & 3,807            & 4,313 \\
    \bottomrule
  \end{tabular}
\end{table}


\end{document}